\documentclass{IOS-Book-Article}

\usepackage{mathptmx}
\usepackage{amsmath}
\usepackage{amssymb}
\usepackage{amsthm}
\usepackage{booktabs}
\usepackage{orcidlink}
\usepackage{xcolor} 
\theoremstyle{plain}
\newtheorem{proposition}{Proposition}
\theoremstyle{definition}
\newtheorem{definition}{Definition}
\newtheorem{example}{Example}

\newcommand{\Ii}{\mathcal{I}}
\newcommand{\Cc}{\mathcal{C}}
\newcommand{\Aa}{\mathcal{A}}
\newcommand{\Ee}{\mathcal{E}}
\newcommand{\Rr}{\mathcal{R}}
\newcommand{\Bb}{\mathcal{B}}
\newcommand{\Ba}[1]{\mathcal{B}_{a}^{#1}}
\newcommand{\Benv}{\mathcal{B}_{\mathtt{env}}}
\newcommand{\pr}[1]{\mathtt{#1}}
\newcommand{\TU}{\mathrm{TotalUtility}}
\newcommand{\gret}{\gamma_{\mathit{ret}}}
\newcommand{\gans}{\gamma_{\mathit{ans}}}
\newcommand{\ok}{\checkmark}
\newcommand{\ko}{$\times$}

\hypersetup{urlcolor=blue!60!black}

\begin{document}
\begin{frontmatter}

\title{Honest but Wrong: Formalizing Retraction in the Ethics of Speech Acts}
\runningtitle{Honest but Wrong: Formalizing Retraction in the Ethics of Speech Acts}

\author{\fnms{Benjamin} \snm{Icard}\,\orcidlink{0009-0005-4530-5646}%
\thanks{Corresponding Author: Benjamin Icard, benjamin.icard@lip6.fr.}},
\author{\fnms{Gauvain} \snm{Bourgne}\,\orcidlink{0000-0002-5104-9352}},\linebreak
\author{\fnms{Jeanne} \snm{Bonnaventure}\,\orcidlink{0009-0003-7188-9230}}
and
\author{\fnms{Jean-Gabriel} \snm{Ganascia}\,\orcidlink{0000-0002-8181-1379}}

\runningauthor{B. Icard et al.}
\address{LIP6, Sorbonne University, CNRS, Paris, France}

\begin{abstract}
A speaker who asserts honestly may later discover that its assertion was
false. Such erroneous truths are not lies, yet they raise the question of
whether, and on which moral grounds, the speaker must retract them. Using the
FIPA-ACL performative \textit{disconfirm}, we extend a belief-based framework that evaluates speech acts under deontologism and
consequentialism to the case of retraction. We prove that the honesty duty of deontologism, which bears
on utterances when they are made, never requires retracting an erroneous
truth, and add a persistent one, which forbids letting the addressee believe
what the speaker no longer believes. On a legal-advice case implemented in Answer Set Programming, we show that deontologism and consequentialism both
require the retraction, but that consequentialism no longer does once it comes
too late to prevent harm.
\end{abstract}

\begin{keyword}
Speech Acts\sep Retraction\sep Honesty\sep Moral Theories\sep Answer Set
Programming
\end{keyword}
\end{frontmatter}

\section{Introduction}
\label{sec:intro}

The ethics of assertion has mainly been studied through lying, that is,
dishonest utterances made with the intent to
deceive~\cite{bok1978lying,carson2010lying}. Yet a speaker may also make an
honest assertion that turns out to be false. Such an assertion, which we call
an \textit{erroneous truth}~\cite{icard2026formalizing}, raises no issue of
honesty when it is made, since the speaker says what it believes. The issue
arises once the speaker realizes its mistake, for instance after being
corrected or learning new information. Must it then retract its assertion, and
on which grounds? \textcolor{black}{The same issue arises for large language
models, which may assert falsehoods.}

This question is addressed in several fields. In philosophy of language,
retraction is treated as a norm of assertion: on MacFarlane's account, an agent
is required to retract an earlier assertion whose content is not true as
assessed from its current context~\cite[p.~108]{macfarlane2014assessment}. In
professional ethics, correction is treated as a duty: under Rule~3.3(a)(1) of
the ABA Model Rules of Professional Conduct, a lawyer must correct a false
statement of material fact or law previously made to a
tribunal~\cite{aba_rule33}. In agent communication, FIPA-ACL~\cite{fipa2002,labrou1999} provides the \textit{disconfirm} performative, by which a sender informs a receiver that a proposition that the receiver believes, or is uncertain about, is false. \textcolor{black}{In logic, lying has been modeled as an action
transforming the addressee's beliefs~\cite{vanditmarsch2014dynamics},
dishonesty in a modal logic of belief and
intention~\cite{sakama2015dishonesty}, deception, including
self-deception~\cite{sakama2015deception}, and persuasion as influencing
choices by changing beliefs~\cite{bonnet2021influencing}.} None of these approaches, however, compares how moral theories evaluate retraction.

In previous work, we proposed a belief-based framework for the moral
evaluation of speech acts~\cite{icard2026formalizing}, implemented in
Answer Set Programming (ASP)~\cite{gelfondlifschitz1991classical}, which assesses utterances under
several moral theories, including deontologism and consequentialism. This framework evaluates
each utterance at the moment it is made, and leaves to future work the
situations in which the agent's beliefs change after it has acted. In this
paper, we address these situations for assertive speech acts, taking the
speaker's later beliefs as given. We make the following contributions:
\begin{itemize}
\item \textit{Retraction.} We extend the framework to two steps with the
  \textit{disconfirm} performative, and characterize a past assertion as
  maintained, retracted or not retracted.
\item \textit{Persistent honesty.} We show that instantaneous honesty never
  condemns not retracting an erroneous truth, and add a persistent honesty
  duty to deontologism.
\item \textit{Evaluation.} We evaluate the extension in ASP on a legal-advice
  case and a variant, showing where deontologism and consequentialism converge
  and diverge.
\end{itemize}
\section{Background}
\label{sec:background}

\paragraph{Vocabulary.} The framework of~\cite{icard2026formalizing} fixes four disjoint sets of ground
atoms: invariants $\Ii$, context atoms $\Cc$, actions $\Aa$ and events $\Ee$.
A \textit{belief base} $\Bb \subseteq \Cc$ interprets the uncertain context,
and a \textit{decision} $D \subseteq \Aa$ is the set of actions selected by an
agent. We write $\Ba{}$ for the belief base of the evaluated agent $a$, and
$\Benv$ for the actual state of affairs. Communicative acts are represented by
the predicate
$\pr{utter}(X,Y,N,\varphi(\sigma))$, where $X$ is the \textit{speaker}, $Y$ the \textit{addressee}, $N$ the
\textit{step} at which it is made, and $\varphi$ a \mbox{FIPA-ACL}
performative~\cite{fipa2002} applied to the \textit{content} $\sigma$, such as
$\pr{inform}(\sigma)$ for an assertion, with $\sigma \in \Cc \cup \Ee$, or
$\pr{query}(\sigma(x))$ for a question, with $x$ free. Utterances of $a$ are
actions, whereas utterances addressed to $a$ are events.

\paragraph{Causal and Decision Theories.} A \textit{causal theory} $\Delta$,
inspired by structural equation models~\cite{pearl2009causality}, is an
acyclic set of rules $\varepsilon \leftarrow \delta$, with
$\varepsilon \in \Ee$ and $\delta$ a conjunction of literals over
$\Ii \cup \Cc \cup \Aa \cup \Ee$. Under a closed-world reading, it determines
which events hold for each pair $(\Bb,D)$, and we write $\sigma_{\Bb,D}$ when
$\sigma$ holds.

A \textit{decision theory} $\Pi$ is a set of rules
$\gamma \leftarrow \delta$ whose heads are \textit{choices} of the form
$\gamma = \langle \Rr, \min, \max \rangle$, where $\Rr \subseteq \Aa$ is a set
of offered actions, and $\min$ and $\max$ are the minimum and maximum numbers of
actions of $\Rr$ that a decision may contain. Given the set
$\Gamma_{\Bb,D} = \{\gamma \mid (\gamma \leftarrow \delta) \in \Pi
\text{ and } \delta_{\Bb,D}\}$ of triggered choices, a decision $D$ is
\textit{admissible} for $\Bb$ iff every action of $D$ is offered by some
$\gamma \in \Gamma_{\Bb,D}$, and
$\gamma.\min \le |\gamma.\Rr \cap D| \le \gamma.\max$ for every
$\gamma \in \Gamma_{\Bb,D}$.

\paragraph{Epistemic Types.}\looseness=-1 An assertion $\pr{utter}(a,q,N,\pr{inform}(\sigma))$
of $a$ to an addressee $q$ is \textit{honest} if $a$ believes $\sigma$, which is
the sincerity condition of $\pr{inform}$, and \textit{dishonest} otherwise:
\begin{align*}
\pr{honest}(a,\sigma) &\leftarrow
  \pr{utter}(a,q,N,\pr{inform}(\sigma)) \wedge \sigma_{\Ba{},D},\\
\pr{dishonest}(a,\sigma) &\leftarrow
  \pr{utter}(a,q,N,\pr{inform}(\sigma)) \wedge \neg\sigma_{\Ba{},D}.
\end{align*}
It is \textit{true} or \textit{false} according to whether $\sigma_{\Benv,D}$
holds. Crossing these two dimensions yields four types: \textit{objective
truths} (honest and true), \textit{erroneous truths} (honest and false),
\textit{objective lies} (dishonest and false) and \textit{erroneous lies}
(dishonest and true).

\paragraph{Moral Theories.} We consider two of the moral theories under which
the framework evaluates decisions. As the causal theory is acyclic, each
decision $D$ determines a single \textit{situation} $S$, and we write
$\mathbb{S}$ for the set of candidate situations.
\begin{itemize}
\item \textit{Deontologism} assesses situations against absolute maxims, each
  violated under a condition $\delta$, written
  $\mathrm{violated}_S(m) \leftarrow \delta$. A situation is permissible if it
  violates no maxim, and impermissible otherwise. We use the unconditional
  honesty duty $\pr{m1}$:
  \begin{align*}
  \mathrm{violated}_S(\pr{m1}) \leftarrow \pr{dishonest}(a,\sigma).
  \end{align*}
\item \textit{Consequentialism} ranks situations by their total utility under
  a belief base $\Bb$, that is, the sum of the utilities of the events
  $\Ee_{S,\Bb}$ of $S$ under $\Bb$, given a utility function
  $\mathrm{utility} : \Ee \to \mathbb{Z}$. A situation $S$ is permissible iff
  \begin{align*}
  S \in \arg\max_{S' \in \mathbb{S}} \TU(S',\Bb),
  \quad\text{with}\quad
  \TU(S,\Bb) = \sum\nolimits_{\varepsilon \in \Ee_{S,\Bb}} \mathrm{utility}(\varepsilon),
  \end{align*}
  and $\Bb = \Ba{}$ for expected utility ($\pr{c1}$) and $\Bb = \Benv$ for
  actual utility ($\pr{c2}$).
\end{itemize}

\section{Retracting Assertions}
\label{sec:retraction}

\paragraph{Two-Step Decisions.} At step~$0$, agent $a$ holds the belief base
$\Ba{0}$ and takes a decision $D_0$ admissible for $\Ba{0}$. At step~$1$, it
holds $\Ba{1}$ and takes a decision $D_1$ admissible for $\Ba{1}$. We treat
$\Ba{1}$ as an input of the model: how it is obtained from $\Ba{0}$ and from
the agent's perceptions lies outside the scope of this paper. Since utterances
cannot be unsaid, we require $D_0 \subseteq D_1$, where $D_1 \setminus D_0$
only contains actions of step~$1$.

\begin{definition}[step-indexed epistemic type]\label{def:type}
Let $u = \pr{utter}(a,q,N,\varphi(\sigma))$ be an utterance made at step
$N \in \{0,1\}$. The honesty of $u$ is evaluated with respect to $\Ba{N}$ and
$D_N$, and its truth with respect to $\Benv$ and $D_N$.
\end{definition}

An erroneous truth thus never turns into a lie afterwards.

\paragraph{The Disconfirm Performative.} In FIPA-ACL, the act
$\langle i, \pr{disconfirm}(j,\phi)\rangle$ has the feasibility precondition
$B_i\neg\phi \wedge B_i(U_j\phi \vee B_j\phi)$, where $B$ and $U$ denote belief
and uncertainty~\cite{fipa2002}. As for \textit{inform} in the framework of \cite{icard2026formalizing}, we do not represent nested beliefs: the second conjunct is
approximated by the past assertion of $\sigma$ by $a$ to $q$, together with the
context atom $\pr{credible}(\sigma,q)$, which states that $q$ would find
$\sigma$ credible. \textcolor{black}{As an atom of $a$'s belief base, it
encodes $a$'s belief about $q$'s attitude, without the inferences that
epistemic logics license on nested beliefs~\cite{vanditmarsch2014dynamics},
which we give up to keep the ASP setting of~\cite{icard2026formalizing}.} By Definition~\ref{def:type},
$\pr{utter}(a,q,N,\pr{disconfirm}(\sigma))$ is honest iff
$\neg\sigma_{\Ba{N},D_N}$. For each addressee $q$ and content $\sigma$, the
decision theory is extended with the following rule and choice:
\begin{align*}
\gret(q,\sigma) &\leftarrow
  \pr{utter}(a,q,0,\pr{inform}(\sigma)) \wedge \pr{credible}(\sigma,q)
  \wedge \neg\sigma,\\
\gret(q,\sigma) &= \langle \{\pr{utter}(a,q,1,\pr{disconfirm}(\sigma))\}, 0, 1 \rangle.
\end{align*}
Its bounds make the retraction
available but not mandatory, leaving its moral status to
Section~\ref{sec:ethics}. For an honest assertion, $\gret(q,\sigma)$ is not
triggered at step~$0$, and is triggered at step~$1$ exactly when $a$ no longer
believes $\sigma$ while still believing that $q$ finds it credible.

\begin{definition}[status of a past assertion]\label{def:status}
Let $\pr{utter}(a,q,0,\pr{inform}(\sigma)) \in D_0$. At step~$1$, this
assertion is \textit{maintained} if $a$ still believes $\sigma$, that is,
$\sigma_{\Ba{1},D_1}$. Otherwise, if $a$ still believes that $q$ finds $\sigma$
credible, so that
$\gret(q,\sigma) \in \Gamma_{\Ba{1},D_1}$, the assertion is \textit{retracted}
when $D_1$ contains $\pr{utter}(a,q,1,\pr{disconfirm}(\sigma))$, and
\textit{not retracted} when it does not.
\end{definition}

An assertion that $q$ would not find credible leaves no belief to correct, and
receives none of these statuses. Combining Definitions~\ref{def:type}
and~\ref{def:status}, an \textit{erroneous truth not retracted} is an honest
assertion that its speaker now disbelieves, but lets its addressee keep
believing.

\begin{example}[appeal deadline]\label{ex:case}
A client $c$ asks a legal assistant $a$ for the deadline to appeal a decision,
that is, the event $\pr{utter}(c,a,0,\pr{query}(\pr{deadline}(\pr{appeal},x)))$. Relying on an
outdated rule, $a$ believes this deadline to be 30 days, whereas it is
actually 15 days, and learns the current rule before step~$1$. With
$\sigma = \pr{deadline}(\pr{appeal},\pr{d30})$:
\begin{align*}
\Ba{0} &= \{\sigma,\ \pr{credible}(\sigma,c)\},\\
\Ba{1} = \Benv &= \{\pr{deadline}(\pr{appeal},\pr{d15}),\
\pr{credible}(\sigma,c)\}.
\end{align*}
At step~$0$, $a$ decides $D_0 = \{\pr{utter}(a,c,0,\pr{inform}(\sigma))\}$,
an erroneous truth, honest under $\Ba{0}$ and false under $\Benv$. At
step~$1$, $\gret(c,\sigma)$ is triggered, since $\sigma$ no longer holds under
$\Ba{1}$ while $\pr{credible}(\sigma,c)$ does. The assertion is not retracted
in $D_1^{\mathit{not\_re}} = D_0$, and retracted in
$D_1^{\mathit{re}} = D_0 \cup \{\pr{utter}(a,c,1,\pr{disconfirm}(\sigma))\}$.
\end{example}

\section{Persistent Honesty}
\label{sec:ethics}

The honesty maxim $\pr{m1}$ recalled in Section~\ref{sec:background} is
\textit{instantaneous}: by Definition~\ref{def:type}, it evaluates each
utterance against the belief base of the step $N$ at which it is made,
and is thus silent on assertions not retracted (Proposition~\ref{prop:m1}). We
introduce a duty of \textit{persistent honesty}: an agent should retract an
assertion it no longer believes. At step~$1$, we compare the
extensions of $D_0$ admissible for $\Ba{1}$.

\paragraph{Deontologism.} We add a maxim $\pr{m2}$, violated by any assertion that
$a$ no longer believes but does not retract, written $\pr{not\_re}(a,\sigma)$
(Definition~\ref{def:status}):
\begin{align*}
\mathrm{violated}_S(\pr{m2}) \leftarrow \pr{not\_re}(a,\sigma).
\end{align*}

\paragraph{Consequentialism.} The theory is unchanged: $\pr{c1}$ and $\pr{c2}$
compare the candidate situations by expected and actual utility, under
$\Ba{1}$ and $\Benv$.

\begin{proposition}[instantaneous honesty]\label{prop:m1}
Assume that $\pr{disconfirm}$ utterances are only offered by retraction
choices. Let all utterances of $D_0$ be honest, and let $D_1$ be admissible for
$\Ba{1}$, with $D_1 \setminus D_0$ consisting of $\pr{disconfirm}$ utterances
only. Then $\pr{m1}$ is not violated in $D_1$. In particular, it permits
not retracting an erroneous truth.
\end{proposition}

\begin{proof}
By Definition~\ref{def:type}, the utterances of $D_0$ keep their step-$0$
type, and are thus honest. Let
$u = \pr{utter}(a,q,1,\pr{disconfirm}(\sigma)) \in D_1 \setminus D_0$. As
$D_1$ is admissible for $\Ba{1}$, $u$ is offered by some choice of
$\Gamma_{\Ba{1},D_1}$, which can only be $\gret(q,\sigma)$, whose condition
requires $\neg\sigma_{\Ba{1},D_1}$. Hence $u$ is honest by
Definition~\ref{def:type}, and $D_1$ contains no dishonest utterance, the only
violation condition of $\pr{m1}$. Taking $D_1 = D_0$ yields the last claim.
\end{proof}

\section{Case Study and Implementation}
\label{sec:case}

\paragraph{Theories.} We complete Example~\ref{ex:case} with the causal theory
$\Delta$:
\begin{align*}
&\pr{corrected}(c,\sigma) \leftarrow
  \pr{utter}(a,c,N,\pr{disconfirm}(\sigma)) \wedge \neg\pr{too\_late},\\
&\pr{relies}(c,\sigma) \leftarrow
  \pr{utter}(a,c,N,\pr{inform}(\sigma)) \wedge \pr{credible}(\sigma,c)
  \wedge \neg\pr{corrected}(c,\sigma),\\
&\pr{harm}(c,\pr{missedDeadline}) \leftarrow
  \pr{relies}(c,\pr{deadline}(\pr{appeal},L))\\
&\qquad \wedge \pr{deadline}(\pr{appeal},L') \wedge \pr{shorter}(L',L),
\end{align*}
with the invariant $\pr{shorter}(\pr{d15},\pr{d30})$ and a utility of $-3$ for
harm events. Besides $\gret$, a choice
$\gans$, triggered by the query, requires $a$ to either inform $c$ of one
candidate deadline or remain silent. In the \textit{too-late} variant, the
period has expired when $a$ learns the rule: $\pr{too\_late}$ is added to
$\Ba{1}$ and $\Benv$, with no rule change.

\begin{table}[t]
\caption{Moral evaluation at step~$1$ of Example~\ref{ex:case} (Base) and its
too-late variant (Too late). \textbf{m1} and \textbf{m2} are the
instantaneous and persistent honesty maxims, \textbf{c1} and \textbf{c2} the
expected- and actual-utility consequentialisms, with \ok: permissible, \ko:
impermissible.}
\label{tab:results}
\centering
\begin{tabular}{@{}llcccc@{}}
\toprule
Scenario & Decision & \textbf{m1} & \textbf{m2} & \textbf{c1} & \textbf{c2}\\
\midrule
Base     & not retract ($D_1^{\mathit{not\_re}}$) & \ok & \ko & \ko & \ko\\
         & retract ($D_1^{\mathit{re}}$)        & \ok & \ok & \ok & \ok\\
\midrule
Too late & not retract ($D_1^{\mathit{not\_re}}$) & \ok & \ko & \ok & \ok\\
         & retract ($D_1^{\mathit{re}}$)        & \ok & \ok & \ok & \ok\\
\bottomrule
\end{tabular}
\end{table}

\paragraph{Results.} We implement the extension in ASP with Python~3.13 and
clingo~5.8.2~\cite{gebser2019multi}. The code is available at
\url{https://github.com/benjaminicard/jurix2026}.
As Table~\ref{tab:results} shows, retracting is permissible under deontologism
and consequentialism in both scenarios. On the one hand, not retracting is always permitted by
the original maxim $\pr{m1}$ (Proposition~\ref{prop:m1}), and always forbidden
by the persistent maxim $\pr{m2}$. On the other hand, consequentialism forbids it in the base
scenario, where the client would miss the deadline, but permits it in the
too-late variant, where harm is unavoidable.

\section{Conclusion}
\label{sec:conclusion}

We extended a belief-based framework for the moral evaluation of speech acts \cite{icard2026formalizing} to the case of retraction, modeled with the FIPA-ACL \textit{disconfirm}
performative. Since the original honesty maxim of deontologism never requires
a retraction, we introduced a persistent honesty duty. On a legal-advice case,
deontologism and consequentialism converge while a retraction prevents harm,
but diverge once it no longer does.

Future work will pursue \textcolor{black}{four} directions: 1)~deriving the speaker's later
beliefs from its observations, through observation-based revision, over
interactions of arbitrary length; 2)~modeling observation actions, so as to
capture duties to inform oneself before asserting or before letting an
assertion stand; 3)~covering further performatives, notably commissives
such as promises, whose withdrawal raises similar moral questions;
\textcolor{black}{and 4)~representing nested beliefs, as in dynamic epistemic
logic~\cite{vanditmarsch2014dynamics}.}

\paragraph{Acknowledgements.} We thank three anonymous reviewers for their helpful comments and suggestions. BI, GB and JGG acknowledge the programs THEMIS (grant agreements DOS022279400 and DOS022279500), TRUSTEDNEWS (ANR-25-ASM2-0003) and VIGILANT (ANR-26-ASRE-0004). The authors used Claude Opus~5.5 \mbox{(Anthropic)} solely to edit the text and correct its phrasing, and take full responsibility for the manuscript, whose ideas, writing and content are their own.

\bibliographystyle{vancouver}
\bibliography{refs}

@inproceedings{icard2026formalizing,
  author    = {Benjamin Icard and Gauvain Bourgne and Jeanne Bonnaventure and Jean-Gabriel Ganascia},
  title     = {Formalizing the Moral Evaluation of Speech Acts: Truthfulness, Lies and Ethical Dilemmas},
  booktitle = {PRIMA 2026: Principles and Practice of Multi-Agent Systems},
  publisher = {Springer Nature Switzerland},
  year      = {2026},
  note      = {To appear},
  url = {https://arxiv.org/abs/2610.04747}
}

@misc{fipa2002,
  author       = {{FIPA}},
  title        = {{FIPA} Communicative Act Library Specification},
  howpublished = {FIPA Standard SC00037J, Foundation for Intelligent Physical Agents},
  year         = {2002},
  url          = {http://www.fipa.org/specs/fipa00037/}
}

@article{labrou1999,
  author  = {Labrou, Yannis and Finin, Tim and Peng, Yun},
  title   = {Agent Communication Languages: The Current Landscape},
  journal = {IEEE Intelligent Systems},
  volume  = {14},
  number  = {2},
  pages   = {45--52},
  year    = {1999}
}

@book{macfarlane2014assessment,
  author    = {MacFarlane, John},
  title     = {Assessment Sensitivity: Relative Truth and Its Applications},
  publisher = {Oxford University Press},
  address   = {Oxford},
  year      = {2014}
}

@misc{aba_rule33,
  author       = {{American Bar Association}},
  title        = {Model Rules of Professional Conduct},
  howpublished = {Rule 3.3: Candor Toward the Tribunal (adopted 1983, as amended)},
  year         = {1983}
}

@book{pearl2009causality,
  author    = {Pearl, Judea},
  title     = {Causality: Models, Reasoning, and Inference},
  edition   = {2nd},
  publisher = {Cambridge University Press},
  address   = {Cambridge},
  year      = {2009}
}

@article{gelfondlifschitz1991classical,
  author  = {Gelfond, Michael and Lifschitz, Vladimir},
  title   = {Classical Negation in Logic Programs and Disjunctive Databases},
  journal = {New Generation Computing},
  volume  = {9},
  number  = {3--4},
  pages   = {365--385},
  year    = {1991}
}

@article{gebser2019multi,
  author  = {Gebser, Martin and Kaminski, Roland and Kaufmann, Benjamin and Schaub, Torsten},
  title   = {Multi-shot {ASP} Solving with clingo},
  journal = {Theory and Practice of Logic Programming},
  volume  = {19},
  number  = {1},
  pages   = {27--82},
  year    = {2019}
}

@book{bok1978lying,
  author    = {Bok, Sissela},
  title     = {Lying: Moral Choice in Public and Private Life},
  publisher = {Pantheon Books},
  address   = {New York},
  year      = {1978}
}

@book{carson2010lying,
  author    = {Carson, Thomas L.},
  title     = {Lying and Deception: Theory and Practice},
  publisher = {Oxford University Press},
  address   = {Oxford},
  year      = {2010}
}

@article{vanditmarsch2014dynamics,
  author  = {van Ditmarsch, Hans},
  title   = {Dynamics of Lying},
  journal = {Synthese},
  volume  = {191},
  number  = {5},
  pages   = {745--777},
  year    = {2014}
}

@article{sakama2015dishonesty,
  author  = {Sakama, Chiaki and Caminada, Martin and Herzig, Andreas},
  title   = {A Formal Account of Dishonesty},
  journal = {Logic Journal of the IGPL},
  volume  = {23},
  number  = {2},
  pages   = {259--294},
  year    = {2015}
}

@inproceedings{sakama2015deception,
  author    = {Sakama, Chiaki},
  title     = {A Formal Account of Deception},
  booktitle = {Deceptive and Counter-Deceptive Machines: Papers from the 2015 AAAI Fall Symposium},
  publisher = {AAAI Press},
  pages     = {34--41},
  year      = {2015}
}

@inproceedings{bonnet2021influencing,
  author    = {Bonnet, Gr{\'e}gory and Leturc, Christopher and Lorini, Emiliano and Sartor, Giovanni},
  title     = {Influencing Choices by Changing Beliefs: A Logical Theory of Influence, Persuasion, and Deception},
  booktitle = {Deceptive AI. DeceptECAI 2020, DeceptAI 2021},
  editor    = {Sarkadi, Stefan and Wright, Benjamin and Masters, Peta and McBurney, Peter},
  series    = {Communications in Computer and Information Science},
  volume    = {1296},
  publisher = {Springer},
  address   = {Cham},
  pages     = {124--141},
  year      = {2021}
}

\end{document}